\documentclass[pdflatex,sn-mathphys-num]{sn-jnl}

\usepackage{latexsym}
\usepackage{graphicx}%
\usepackage{multirow}%
\usepackage{amsmath,amssymb,amsfonts}%
\usepackage{amsthm}%
\usepackage{mathrsfs}%
\usepackage[title]{appendix}%
\usepackage{xcolor}%
\usepackage{textcomp}%
\usepackage{manyfoot}%
\usepackage{booktabs}%
\usepackage{algorithm}%
\usepackage{algorithmicx}%
\usepackage{algpseudocode}%
\usepackage{listings}%
\usepackage{url}
\usepackage{siunitx}
\usepackage{diagbox}
\usepackage{ifthen}

\theoremstyle{thmstyleone}%

\theoremstyle{thmstyletwo}%

\theoremstyle{thmstylethree}%
\newtheorem{definition}{Definition}

\newboolean{showcomments}
\setboolean{showcomments}{false} 
\ifthenelse{\boolean{showcomments}}
{\newcommand{\nb}[2]{
		\fcolorbox{black}{yellow}{\bfseries\sffamily\scriptsize#1}
		{\sf\small$\blacktriangleright$\textit{#2}$\blacktriangleleft$}
	}
	
}
{\newcommand{\nb}[2]{}
	
}

\newcommand{\rev}[1]{#1}

\begin{document}

\title[Graph Contextual Reinforcement Learning]{Graph Contextual Reinforcement Learning for Efficient Directed Controller Synthesis}

\author[1]{\fnm{Toshihide} \sur{Ubukata}}
\author[2]{\fnm{Enhong} \sur{Mu}}
\author[1]{\fnm{Takuto} \sur{Yamauchi}}
\author[2]{\fnm{Mingyue} \sur{Zhang}}
\author*[1,3]{\fnm{Jialong} \sur{Li}}\email{lijialong@fuji.waseda.jp}
\author[3]{\fnm{Kenji} \sur{Tei}}

\affil[1]{\orgname{Waseda University}, \orgaddress{\city{Tokyo}, \postcode{169-8050}, \country{Japan}}}
\affil[2]{\orgname{Southwest University}, \orgaddress{\country{China}}}
\affil[3]{\orgname{Institute of Science Tokyo}, \orgaddress{\city{Tokyo}, \postcode{152-8550}, \country{Japan}}}

\abstract{Controller synthesis is a formal method approach for automatically generating Labeled Transition System (LTS) controllers that satisfy specified properties. The efficiency of the synthesis process, however, is critically dependent on exploration policies. These policies often rely on fixed rules or strategies learned through reinforcement learning (RL) that consider only a limited set of current features. To address this limitation, this paper introduces GCRL, an approach that enhances RL-based methods by integrating Graph Neural Networks (GNNs). GCRL encodes the history of LTS exploration into a graph structure, allowing it to capture a broader, non-current-based context. In a comparative experiment against state-of-the-art methods, GCRL exhibited superior learning efficiency and generalization across four out of five benchmark domains, except one particular domain characterized by high symmetry and strictly local interactions.}


\maketitle

\section{Introduction}
\label{sec:intro}

In modern software engineering, it is essential that complex systems are not only functional but also provably correct. This is especially critical in safety-sensitive domains such as aerospace \cite{SESoS24} and railway systems, where failures can have severe consequences. Controller synthesis \cite{Pnueli, Control_Discovery} is a key formal method that addresses this challenge by automatically generating a controller—typically represented as a Labeled Transition System (LTS)—that is guaranteed to satisfy specified properties, such as safety, with respect to a given model of its environment. The appeal of this approach lies in its ability to produce correct-by-construction systems, automating a particularly challenging aspect of system design \cite{Sardina2015}.

Despite its advantages, a major barrier to the practical adoption of controller synthesis is the state-space explosion problem. The total number of system states can grow exponentially with the number of components and the complexity of the specifications, making it infeasible to construct and explore the entire state space. To mitigate this issue, on-the-fly Directed Controller Synthesis (DCS) \cite{CiolekDCS} offers a promising alternative. Rather than constructing the full state space in advance, DCS incrementally explores only the relevant portions needed to synthesize a correct controller, thereby managing the exponential growth more effectively.

The effectiveness of DCS, however, hinges on the design of its exploration policy—the strategy that determines which frontier state to examine next. A well-designed exploration policy acts as an effective heuristic, efficiently steering the search toward promising regions of the state space and away from unproductive ones, thus reducing synthesis costs. Conversely, a poor policy may direct the search into irrelevant areas, wasting time and computational resources. To improve exploration strategies, recent research has applied Reinforcement Learning (RL) to automatically learn effective policies \cite{Delgado_Sanchez}. By framing policy design as an RL problem, a learning agent can discover strategies that outperform manually crafted heuristics.

However, current RL-based approaches suffer from a critical limitation: they base decisions almost exclusively on local features of immediate successor states. This is analogous to an explorer navigating a forest using only what’s visible in front of them—able to evaluate nearby terrain but unaware of the broader trails already explored or where they might lead. As a result, the RL agent lacks contextual awareness and cannot leverage information from its exploration history, such as recurring structural patterns or early indicators of dead ends. This contextual blindness significantly limits the agent’s ability to learn informed, forward-looking DCS exploration policies.

To overcome this limitation, we propose Graph Contextual Reinforcement Learning (GCRL)—an approach that incorporates structural information from the exploration history into the decision-making process. The core idea of GCRL is to model the already explored portion of the LTS as a graph at each decision point, and then encode and process this graph using Graph Neural Networks (GNNs) \cite{GNN}. GNNs can aggregate information across the graph, producing rich node embeddings that capture both local and global relationships, as well as structural patterns over time. This gives the RL agent a bird’s-eye view of the explored space, enabling it to make decisions informed by both current and historical context.

The main contributions of this paper are:
\begin{itemize}
\item We introduce a way to integrate GNNs into the reinforcement learning framework for controller synthesis, enabling the policy to utilize historical exploration context.
\item We propose a method for dynamically building a graph that captures the changing structure of the explored LTS, including key features of both past and potential next steps.
\item We evaluate our approach on five benchmarks and show that GCRL learns faster and its policies perform better zero-shot generalization on larger, unseen problems compared to existing RL methods.
\end{itemize}

The initial idea for this work was first presented in a conference abstract paper~\cite{ubukata2025graph}. In this extended work, we have made the following significant improvements:
(1) In terms of technical improvements, we enhanced computational efficiency by refining the GNN's Q-value estimation for frontier edges. This involves k-hop subgraph pruning and focused Q-value computation, ensuring the GNN processes only the most relevant local exploration context.
(2) For the evaluation part, we have significantly expanded our empirical validation by testing GCRL on five diverse benchmark problem domains instead of just one, to demonstrate its broader applicability. In addition, we conducted a more in-depth discussion and analysis of the experimental results.

The rest of this paper is organized as follows: Section~\ref{sec:background} gives the necessary background. Section~\ref{sec:proposal} explains our GCRL approach in detail. Section~\ref{sec:evaluation} presents and discusses the results. Section~\ref{sec:related} reviews related work. Finally, Section~\ref{sec:conclusion} summarizes the paper and discusses future work.

\section{Background}
\label{sec:background}

This section outlines the theoretical foundations for our approach. We begin by formalizing the concepts of Discrete Event Systems (DES) and Directed Controller Synthesis (DCS). We then define the specific optimization problem that arises from on-the-fly synthesis methods. Finally, we detail a state-of-the-art reinforcement learning approach for solving this problem, which serves as the primary baseline for this study.

\subsection{Directed Controller Synthesis}

\begin{definition}[Discrete Event System]
A DES is a 5-tuple $E = (S_E, A_E, D_E, s_0, M_E)$, where:
\begin{itemize}
    \item $S_E$ is a finite set of states.
    \item $A_E$ is a finite set of event labels, partitioned into controllable events $A_E^C$ and uncontrollable events $A_E^U$, such that $A_E = A_E^C \cup A_E^U$.
    \item $D_E \rev{\subseteq} S_E \times A_E \times S_E$ is a partial state transition function.
    \item $s_0 \in S_E$ is the initial state.
    \item $M_E \subseteq S_E$ is a set of marked states, which typically represent the completion of a task.
\end{itemize}
\end{definition}

In supervisory control theory, the system to be controlled is modeled as a DES, typically as defined in the modular framework of Ramadge and Wonham~\cite{Wonham1988}. Here, an automaton $E$ defines a language $Lg(E) \subseteq A_E^*$, where $^*$ denotes the Kleene closure. A word $w \in A_E^*$ belongs to the language if it corresponds to a valid path of transitions from the initial state. We denote a transition path from $s_0$ to $s_t$ on word $w$ as $s_0 \xrightarrow{w}_E s_t$. A controller's role is to restrict the behavior of the plant by disabling controllable events.
\rev{As DES is a well-established concept in control and computer science, a detailed introduction is omitted here. Interested readers can refer to \cite{SILVA2018213} for detailed background.
}

\begin{definition}[Controller and Director]
Given a DES $E$, a controller is a function $\sigma: A_E^* \to \mathcal{P}(A_E^C)$ that maps an observed event sequence to a set of permissible controllable events. The language generated by the controlled system, denoted $Lg^\sigma(E)$, consists of all words $w = l_0 \ldots l_{k-1} \in Lg(E)$ such that for each event $l_i$, either $l_i \in A_E^U$ (it is uncontrollable) or $l_i \in \sigma(l_0 \ldots l_{i-1})$ (it is enabled by the controller).
A controller $\sigma$ is non-blocking if for any trace $w \in Lg^\sigma(E)$, there exists a non-empty continuation $w' \in A_E^*$ such that the concatenated word $ww' \in Lg^\sigma(E)$ leads to a marked state. Furthermore, a controller is a director if it is maximally permissive, allowing at most one controllable event at any time, i.e., $|\sigma(w)| \leq 1$ for all $w \in A_E^*$ \cite{Huang:2008:DCD}.
\end{definition}

A non-blocking controller is inherently safe, meaning it prevents the system from reaching a deadlock state (a state with no outgoing transitions). Complex systems are often modeled in a modular fashion using the parallel composition of multiple automata \cite{Ramadge:1989:SC}, which synchronizes on shared events and interleaves on private ones. The parallel composition of two DES $T$ and $Q$ yields $T \| Q = (S_T \times S_Q, A_T \cup A_Q, D_{T\|Q}, \langle t_0,q_0 \rangle, M_T \times M_Q)$.

A modular directed controller synthesis problem is thus defined by a set of automata $(E^1, \ldots, E^n)$, and its solution is a non-blocking director for the composed plant $E^1\|\ldots\|E^n$.

\subsection{Exploration Optimization in On-the-Fly Synthesis}
Solving a modular directed control problem by first explicitly constructing the full state space of the composed plant is often intractable due to the state-space explosion problem. On-the-fly approaches, such as DCS \cite{CiolekDCS}, aim to mitigate this by exploring the state space incrementally. These methods search for a valid control strategy (or prove its non-existence) by analyzing only a small subset of the plant.

The core challenge of on-the-fly synthesis is guiding the exploration process efficiently. The process builds an exploration sequence, which is the history of explored transitions $h$. This sequence defines the boundary of the search, known as the exploration frontier.

\begin{definition}[Exploration Frontier]
Given a plant $E$ and an exploration sequence $h \subseteq \rev{D_E}$, the exploration frontier $F(E, h)$ is the set of all unexplored transitions $(s, l, s') \in (\rev{D_E} \setminus h)$ whose source state $s$ has been discovered. A state is considered discovered if it is the initial state $s_0$ or if it is the target of a transition in $h$.
\end{definition}

The sequence of choices from the frontier directly impacts the size of the subgraph that must be constructed. This gives rise to an optimization problem centered on the exploration policy.

\begin{definition}[Exploration Optimization Problem]
Given a control problem $E$, the associated exploration optimization problem is to find a heuristic or policy $H$ that selects a transition from the exploration frontier $F(E, h)$ at each step to minimize the total number of transitions in the final sequence $h$ required to find a solution.
\end{definition}

\subsection{Learning Exploration Policies via RL}
A recent approach by \cite{Delgado_Sanchez} frames the exploration optimization problem as a Reinforcement Learning (RL) \cite{Sutton_RL} task to automatically learn a high-performance exploration policy. This method serves as the baseline for this paper.

\subsubsection{MDP Formulation for Exploration}
The core idea is to model the DCS synthesis process itself as a Markov Decision Process (MDP), where the optimal policy for this MDP directly corresponds to an optimal exploration heuristic.

\begin{itemize}
    \item State ($s$): A state in the MDP corresponds to the current exploration history, i.e., the partially explored plant graph defined by the sequence of expanded transitions $h$.
    \item Action ($a$): An action is the selection of a single transition $(s_{des}, \l)$ from the current exploration frontier $F(E, h)$.
    \item Reward ($r$): A constant negative reward of $-1$ is given for every action (i.e., every transition expansion). This incentivizes the agent to find a solution in the fewest possible steps.
    \item Episode Termination: An episode terminates when the DCS algorithm classifies the initial state of the plant as either definitively winning or losing.
\end{itemize}

\subsubsection{State Abstraction and Function Approximation}
This MDP formulation is impractical for direct application due to its high-dimensional, graph-structured state space and a large, variable-sized action space. To overcome this, \rev{the prior RL method \cite{Delgado_Sanchez} introduces} a crucial abstraction layer. Instead of processing the raw graph state and action, it defines a feature-extraction function $\phi(h, a)$ that maps a state-action pair to a fixed-size feature vector. This vector encodes properties of the candidate transition $a = (s_{des}, \l, s'_{des})$ and the current exploration context $h$. Features include the event label type, whether the source/target states are marked, the current classification of the source state (winning/losing/undecided), and other structural properties of the explored graph.

With this abstraction, a modified DQN-style algorithm is used. The Q-network, $Q_\theta$, does not have one output per action. Instead, it takes a feature vector $\phi(h, a)$ as input and produces a single scalar output representing the estimated Q-value for that specific action. At each decision step, the policy evaluates all actions $a_i \in F(E, h)$ by feeding their corresponding feature vectors $\phi(h, a_i)$ into the network and greedily selects the action with the highest Q-value. Standard techniques like experience replay and a fixed target network are employed to ensure stable training.

\subsubsection{Generalization for Exploration Policy}
\label{sec:Generalization}
Another key aspect of the method is its approach to generalizing the exploration policy—that is, it trains a policy on small, tractable problem instances and effectively generalizes it to larger, computationally intensive instances within the same domain. This is achieved through a three-step methodology: (1) Training: The reinforcement learning agent is trained on a small, fixed-size instance of the problem domain (e.g., size $(n=2, k=2)$). During this training process, multiple snapshots of the policy network’s weights are saved at regular intervals. (2) Selection for Generalization: The saved policy snapshots are then evaluated on a set of intermediate-sized instances under a limited time or expansion budget. The policy that achieves the best overall performance across these instances—measured by the number of problems it successfully solves—is selected. This step is designed to explicitly favor generalization capability rather than mere performance on the original training instance. (3) Deployment: Finally, the single policy identified as best at generalizing is used to solve the largest instances in the domain.

\section{Graph Contextual Reinforcement Learning for Learning Exploration Policy}
\label{sec:proposal}

  \begin{figure*}[t] 
    \centering
  \includegraphics[width=\textwidth]{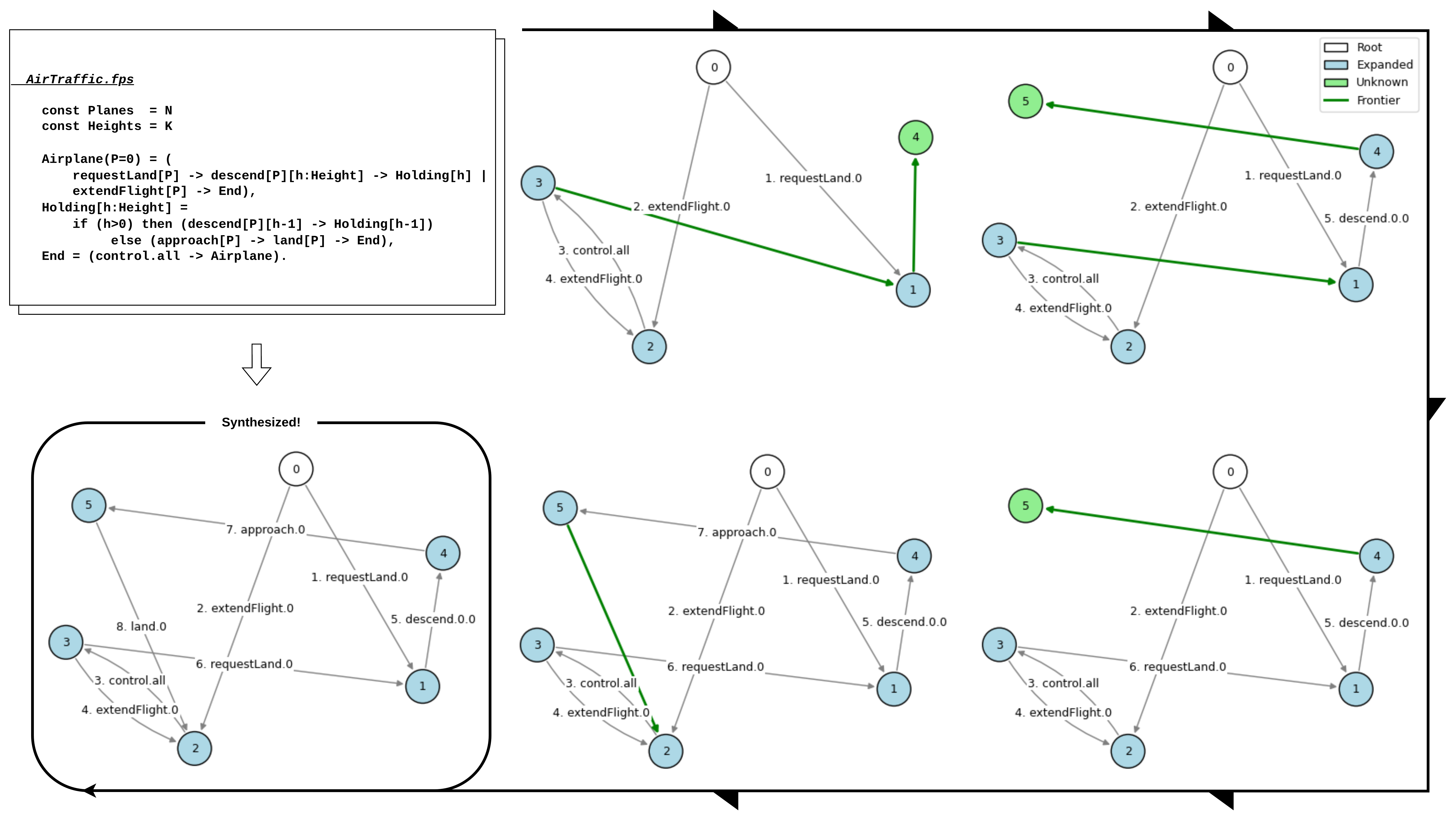} 
    \caption{Step-by-step illustration of controller synthesis in the Air Traffic domain $(n,k)=(1,1)$}
    \label{fig:controller} 
  \end{figure*}

The core idea of GCRL is to use a graph to encode structural and historical exploration context (e.g., explored LTS information), thereby enabling a more efficient RL policy compared to approaches relying solely on current state features. 
In this section, we first outline the graph construction process, followed by the GNN architecture used.
Note that while the state space is originally represented as LTS (but not a standard graph), we dynamically construct a graph encoding from the exploration history of the LTS to feed into the GNN model.

\subsection{Graph Encoding of Contextual Exploration State}
\label{subsec:graph_modeling}
We dynamically construct a graph at each decision point in the DCS process. This graph is rooted at the most recently expanded state and includes nodes and edges encoding previously explored states/transitions (history) as well as the immediately available transitions (frontier).
Algorithm~\ref{alg:graph_construction} outlines this procedure, taking current observations and history information to build the graph structure and associated features, specifically identifying frontier edges available for selection.

\begin{algorithm}[thb]
    \small
    \caption{Graph Encoding Construction}
    \label{alg:graph_construction}
    \begin{algorithmic}[1]

    \State \rev{\textbf{Input:}} Transition history $\mathcal{H}$, Frontiers $\mathcal{F}$
    \State \rev{\textbf{Output:}} Edge indices $\mathcal{E}$, Edge features $\mathcal{A}$, Frontier edge indices $\mathcal{I}$
    
    \State Initialize $\mathcal{E} \gets \emptyset$, $\mathcal{A} \gets \emptyset$, $\mathcal{I} \gets \emptyset$, $u \gets 0$
    
    \Statex \textbf{Step 1: Add transition history}
    \ForAll{transition $(\text{from}, \text{to}, \phi)$ in $\mathcal{H}$}
        \State Append $(\text{from}, \text{to})$ to $\mathcal{E}$; Append $\phi$ to $\mathcal{A}$
    \EndFor
    
    \Statex \textbf{Step 2: Add frontiers}
    \For{$i = 1$ to $|\mathcal{F}|$}
        \State Let $(\text{from}, \text{to}, \phi)$ be the $i$-th frontier
        \If{$\text{to}$ is unknown}
            \State $\text{to} \gets$ new node ID based on $u$; Increment $u$
        \EndIf
        \State Append $(\text{from}, \text{to})$ to $\mathcal{E}$; Append $\phi$ to $\mathcal{A}$
        \State Append $|\mathcal{E}|-1$ to $\mathcal{I}$
    \EndFor
    
    \State \Return $(\mathcal{E}, \mathcal{A}, \mathcal{I})$
    
    \end{algorithmic}
\end{algorithm}

Specifically, nodes and edges are annotated with features capturing state status and transition properties. Node features include (1) just explored flag, (2) explored ratio, (3) uncontrollable flag, (4) marked flag, (5) exploration context/phase.
Edge features include (i) event label, (ii) path labels (summarized), (iii) controllable flag, (iv) leads to marked state flag, (v) synthesis phase info, (vi) child state classification, (vii) child has uncontrollable flag, (viii) child explored status, (ix) source is last expanded flag. 
For readability, we omit the detailed explanation. Interested readers may refer to the Appendix.

Fig.~\ref{fig:controller} illustrates the step-by-step process of controller synthesis using our graph-based encoding.
Given an FSP specification as input, the synthesis proceeds by expanding one frontier transition per step, gradually constructing the controller.
At each step, the graph is updated to reflect the current exploration history and available frontier transitions, as described in Algorithm~\ref{alg:graph_construction}.
This graph includes nodes for previously explored states and newly discovered ones, and edges representing both historical transitions and frontiers.
In the figure, green edges indicate the current frontier transitions (corresponding to those indexed in set $I$ in Algorithm~\ref{alg:graph_construction}), blue nodes represent previously explored states already included in the history $H$, and green nodes correspond to newly discovered (yet unexplored) states added in Step~2 of the algorithm.
Specifically, for each candidate frontier transition (green edge), a subgraph centered on its endpoints is extracted, and Q-values are predicted based on node embeddings and edge features.

\subsection{GNN Architecture for Encoding Graph}
\label{subsec:gnn_rl_arch}
As summarized in Table~\ref{tab:model_summary}, we employ a GNN model to encode and process the constructed graph. Graph convolution layers (\texttt{GCNConv}) propagate information across the graph, allowing the model to learn node embeddings that reflect the current neighborhood and broader exploration history (paths, cycles). An Edge MLP then uses the embeddings of the source and target nodes for each \textit{frontier} edge, along with the edge's own features, to compute a priority score. These scores guide the RL agent's action selection (i.e., which frontier transition to explore next), enabling decisions based on structural and historical contexts.

\begin{table}[thb]
\centering
\small
\renewcommand{\arraystretch}{1.3}
\caption{Overview of GNN Architecture}
\label{tab:model_summary}
\begin{tabular}{lll}
\hline
\textbf{Layer} & \textbf{Input $\rightarrow$ Output} & \textbf{Description} \\
\hline
\texttt{GCNConv1} & $[N, F_n] \rightarrow [N, H]$ & Graph convolution \\
\texttt{GCNConv2} & $[N, H] \rightarrow [N, H]$ & Graph convolution \\
\texttt{Edge MLP} & $[N, 2H + F_e] \rightarrow [N, 1]$ & Lin$\rightarrow$ReLU$\rightarrow$Lin \\
\hline
\end{tabular}
\renewcommand{\arraystretch}{1.0}
\end{table}

\vspace{-5mm}
{\small
\begin{itemize}
    \item $N$ : Number of frontier edges
    \item $F_n$: Dimension of node features
    \item $F_e$: Dimension of edge features
    \item $H$ : Hidden dimension used in the GNN layers
\end{itemize}
}

\subsection{Q-value Estimation for Frontier Edges}
\label{subsec:q_value_estimation}
To estimate the utility of each candidate transition, we estimate Q-values only for the frontier edges—the transitions currently available for exploration.
Unlike conventional GNN-based reinforcement learning approaches that compute embeddings over the entire graph and evaluate all transitions, our approach introduces two critical optimizations aimed at improving computational efficiency without sacrificing expressiveness. First, to reduce the computational burden of GNN processing, we extract a $k$-hop induced subgraph centered on the endpoints of frontier edges.
This pruning strategy filters out remote parts of the exploration history that are unlikely to impact immediate decision-making, while retaining relevant local structural context.
\rev{Here, $k$ denotes the hop radius of the subgraph expansion, which controls the trade-off between computational cost and contextual coverage. 
A smaller $k$ (e.g., 1–2) limits the receptive field and reduces the GNN workload, while a larger $k$ increases accuracy at the expense of runtime.
}
This is motivated by the observation that exploration in LTSs often exhibits locality—transitions near the current frontier contain sufficient information for effective policy decisions.
Second, since only one transition from the frontier is selected at each decision step, the Q-values are computed solely for these candidate actions. This avoids unnecessary inference over non-frontier transitions and reduces the size of the forward pass in the policy network.

The estimation process is summarized in Algorithm~\ref{alg:frontier_q_estimation}.
First, a subgraph is extracted from the global graph by identifying the union of source and target nodes in the current frontier set, and expanding it to include their $k$-hop neighbors (lines 1-4).
Then, node embeddings are generated via message passing in a GCN (lines 5-13).
Finally, the Q-value for each frontier edge is predicted using an MLP that takes as input the concatenated embeddings of the source and target nodes along with the corresponding edge features (lines 14-19).

\begin{algorithm}[thb]
\small
\caption{Q-value Estimation for Frontier Edges via GNN}
\label{alg:frontier_q_estimation}
\begin{algorithmic}[1]

\State \rev{\textbf{Input:}} Full edge list $\mathcal{E}$, edge features $\mathcal{A}$, node features $\mathcal{X}$, frontier edge indices $\mathcal{I}$, \rev{hop parameter $k$}
\State \rev{\textbf{Output:}} Q-values $Q(\mathcal{I}) \in \mathbb{R}^{|\mathcal{I}|}$

\Statex \textbf{Step 1: Subgraph Extraction}
\State Compute node set $\mathcal{V}_\mathcal{I} \gets$ endpoints of edges $\mathcal{E}[i]$ for $i \in \mathcal{I}$
\State Compute induced $k$-hop subgraph $(\mathcal{V}_k, \mathcal{E}_k)$ containing $\mathcal{V}_\mathcal{I}$
\State Extract node features $\mathcal{X}_k \gets \mathcal{X}$ restricted to $\mathcal{V}_k$
\State Extract edge features $\mathcal{A}_k \gets \mathcal{A}$ and edges $\mathcal{E}_k$ accordingly

\Statex \textbf{Step 2: GNN-based Node Embedding}
\ForAll{$v \in \mathcal{V}_k$}
    \State Initialize $\mathbf{h}_v^{(0)} \gets \mathcal{X}_k[v]$
\EndFor
\For{$l = 1$ to $L$}
    \ForAll{$v \in \mathcal{V}_k$}
        \State $\mathbf{h}_v^{(l)} \gets \mathrm{ReLU}\left(\sum_{u \in \mathcal{N}(v)} W_l \mathbf{h}_u^{(l-1)} + b_l\right)$
    \EndFor
\EndFor
\State Set final embeddings $\mathbf{h}_v \gets \mathbf{h}_v^{(L)}$ for all $v \in \mathcal{V}_k$

\Statex \textbf{Step 3: Q-value Regression for Frontier Edges}
\ForAll{$i \in \mathcal{I}$}
    \State Let $(u, v) \gets \mathcal{E}[i]$, and $\phi_{uv} \gets \mathcal{A}[i]$
    \State Form feature vector $\mathbf{z}_i \gets [\mathbf{h}_u \,\|\, \mathbf{h}_v \,\|\, \phi_{uv}]$
    \State Compute Q-value $Q_i \gets \mathrm{MLP}(\mathbf{z}_i)$
\EndFor

\State \Return $Q(\mathcal{I}) = [Q_i]_{i \in \mathcal{I}}$

\end{algorithmic}
\end{algorithm}

\section{Evaluation}
\label{sec:evaluation}
This section evaluates the effectiveness of our GCRL approach compared to baselines.

\subsection{Research Questions}
Our evaluation aims to answer 2 key research questions. 
\begin{itemize}
    \item \textbf{RQ1. Learning Efficiency} -  How efficiently does the GCRL approach learn an effective exploration policy compared to the baseline RL method during training? 
    \item \textbf{RQ2. Policy Generalization} - How well does the exploration policy learned on small problem instances generalize to larger, unseen problem instances in a zero-shot DCS setting? 
\end{itemize}

\subsection{Experimental Setup}
\textbf{Platform}. All experiments were performed using an extended version of the MTSA tool~\cite{MTSA}, on a machine equipped with an Intel Core i9-9960X CPU (3.10GHz), 64GB RAM, and no GPU.

\textbf{Baselines}.
We compare our GCRL against the following baseline: (1) Ready Abstraction (RA), which is a state-of-the-art domain-independent exploration policy that is designed to prioritize transitions during on-the-fly exploration in a structured, rule-based manner \cite{CiolekDCS}; and (2) RL, which is the original RL-based approach proposed in \cite{Delgado_Sanchez}.

\textbf{Benchmarks}.
We use five classic DCS benchmarks, including Air Traffic (AT), Bidding Workflow (BW), Travel Agency (TA), Dining Philosophers (DP), and Transfer Line (TL).
\rev{
We further note that the Cat and Mouse (CM) domain exhibits a substantially larger number of reachable transitions—around 2,000 according to the statistics reported ~\cite{Delgado_Sanchez}, whereas domains such as AT contain only about 50 transitions.
This extreme transition density causes exponential growth in the constructed subgraph, resulting in out-of-memory (OOM) errors in our graph-based implementation.
In particular, since GCRL explicitly encodes exploration history as graphs, memory consumption grows with the number of transitions.
We plan to address this limitation in future work by improving the memory efficiency of our graph construction and representation.
}

\textbf{Training procedure}.
For RQ1, we train exploration policies on small instances with $(n, k) = (2, 2)$ to enable generalization to larger problems across domains. Training proceeds for 100 episodes per configuration equally for all methods, with each transition expansion penalized by a reward of $-1$.
Each episode terminates once the initial state is classified as either winning or losing by DCS algorithm \cite{CIOLEK2020}.
An $\epsilon$-greedy exploration strategy is used during training, where $\epsilon$ is initially set to 1.0 and decayed linearly to 0.01 over time. 
This high initial value encourages broader exploration in the early phase of learning, which is particularly important in sparse-reward environments where local optima can hinder effective policy acquisition~\cite{Gehring2022}.
At the end of each episode, model weights are saved. This entire training process is repeated independently 5 times to reduce stochastic effects.
\rev{Throughout all experiments, the hop parameter for local subgraph extraction was fixed to $k=2$ to balance the trade-off between contextual coverage and computational efficiency. 
This value was determined empirically through preliminary tests with $k=1, 2,$ and $3$, where $k=1$ often led to insufficient contextual information for stable learning, whereas $k=3$ substantially increased computational cost without noticeable performance improvement. }

\textbf{Generalization of Policies}.
For RQ2, we evaluate five independently trained policies to assess their generalization on larger instances across all domains.
To assess scalability, we evaluate the selected policies on previously unseen, larger instances, following Section~\ref{sec:Generalization}.
Each policy is tested on all $(n,k)$ configurations up to $(15,15)$, using a maximum of 5000 expanded transition budget per instance.
Instance evaluation respects the same neighborhood condition as in the selection phase: only instances whose immediate neighbors have been successfully solved are considered. 
The primary metric is the number of instances solved within the time limit, providing a practical measure of each policy’s generalization capability and effectiveness on large-scale synthesis tasks.

  \subsection{Experiment Results}
  \begin{figure*}[t] 
    \centering
  \includegraphics[width=\textwidth]{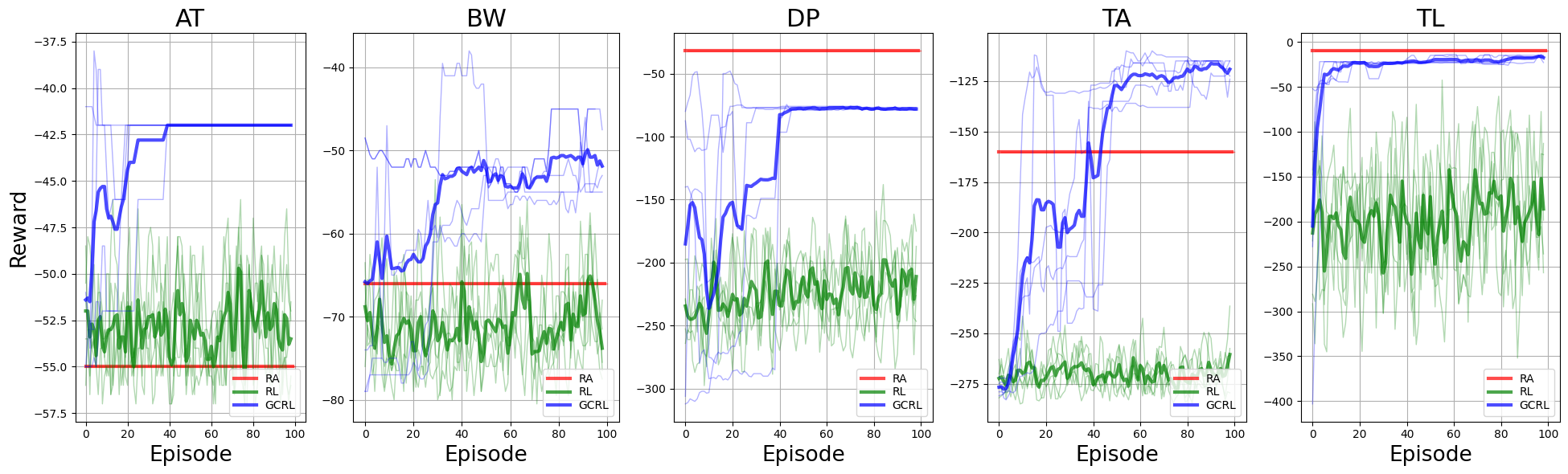} 
    \caption{Comparison of Training Efficiency}
    \label{fig:training} 
  \end{figure*}

  \begin{figure*}[t] 
    \centering
  \includegraphics[width=\textwidth]{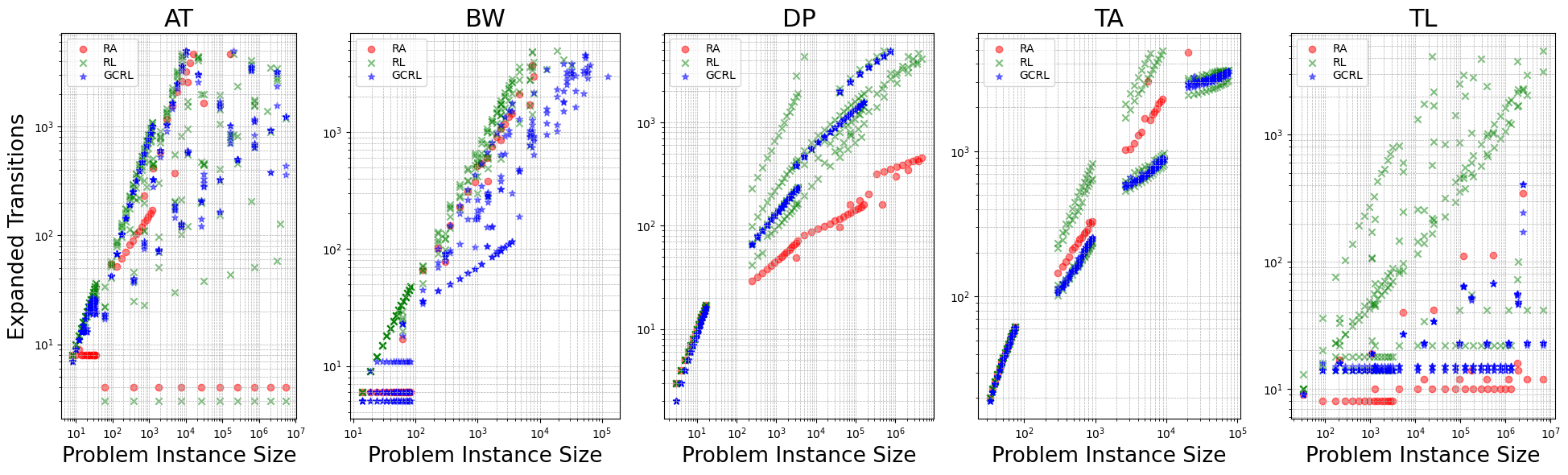} 
    \caption{Comparison of Zero-Shot Generalization}
    \label{fig:generalization}
  \end{figure*}

\begin{table*}[t]
\caption{
Comparison of two evaluation metrics: 
(1) training efficiency measured by the AUC of cumulative negative rewards, 
where \rev{AUC is defined as the absolute area under the cumulative negative reward curve, representing the total exploration cost during training. Smaller AUC values indicate faster learning and higher efficiency}; and 
(2) generalization performance measured by the number of solved instances (higher is better). 
Each value is the mean $\pm$ standard deviation over five runs, except for RA which is fixed.
}
\label{tbl:results}
\centering
\small
\setlength{\tabcolsep}{3pt}
\resizebox{\textwidth}{!}{%
\begin{tabular}{lccccccc}
\toprule
\multirow{2}{*}{\textbf{Domain}} & \multicolumn{3}{c}{\textbf{Training AUC (RQ1)}} & \multicolumn{4}{c}{\textbf{Solved Instances (RQ2)}} \\
\cmidrule(lr){2-4} \cmidrule(lr){5-8}
& RL & GCRL & RL vs GCRL & RA & RL & GCRL & RL vs GCRL \\
\midrule
AT & $-5239.8 \pm 41.4$ & $\mathbf{-4285.30 \pm 105.96}$ & +18.2\% & 57 & $59.6 \pm 15.9$ & $\mathbf{65.2 \pm 0.5}$ & + 9.4\% \\
BW & $-7041.5 \pm 67.3$  & $\mathbf{-5480.10 \pm 463.21}$ & +22.2\% & 38 & $38.2 \pm 2.1$  & $\mathbf{45.6 \pm 2.7}$ & +19.4\% \\
DP & $-22228.9 \pm 556.7$ & $\mathbf{-11150.60 \pm 2947.23}$ & +49.8\% & \textbf{97} & $51.2 \pm 11.6$ & $52.0 \pm 0.0$ & + 1.6\% \\
TA & $-26736.2 \pm 100.0$ & $\mathbf{-15908.20 \pm 1484.5}$ & +40.5\% & 45 & $52.4 \pm 10.6$ & $\mathbf{60.0 \pm 0.0}$ & +14.5\% \\
TL & $-19575.7 \pm 1210.4$ & $\mathbf{-2679.70 \pm 575.75}$ & +86.3\% & 195 & $45.4 \pm 18.8$ & $\mathbf{221.0 \pm 8.9}$ & +386.8\% \\
\bottomrule
\end{tabular}%
}

\end{table*}

\subsubsection{Results on RQ1 - Learning Efficiency}

Figure~\ref{fig:training} illustrates the average number of expanded transitions per episode over 100 training episodes.
Compared to the baseline RL approach, the learning curve of GCRL converges significantly faster across all benchmark domains.
This trend is quantitatively supported by the \rev{Area Under the Curve (AUC)} values summarized in Table~\ref{tbl:results}.
\rev{Here, AUC is defined as the absolute area under the cumulative negative reward curve, representing the total exploration cost during training.}
\rev{Smaller AUC values therefore indicate faster and more efficient learning, as they correspond to lower cumulative exploration cost.}

For example, in the TL domain, GCRL reduces the AUC from $-19575.7$ (RL) to $-2679.7$, indicating an 86.3\% improvement in learning efficiency.

Moreover, we observe that the baseline RL approach has not fully converged after 100 training episodes, as shown in Figure~\ref{fig:training}.
In contrast, GCRL exhibits stable convergence within this episode budget, demonstrating that it can effectively learn an exploration policy even in a limited training regime.
This highlights not only the faster learning speed of GCRL but also its practical applicability in scenarios where training time or data are constrained.

\subsubsection{Results on RQ2 - Policy Generalization}

Figure~\ref{fig:generalization} evaluates the generalization capability of policies trained on small instances $(n,k) = (2,2)$ and tested in a zero-shot manner on larger instances up to $(15,15)$, using a fixed expansion budget.
The $x$-axis represents the size of expanded transitions needed when using monolithic synthesis, while the $y$-axis shows the number of transitions used by each policy (RA, RL, GCRL).
Ideal policies appear in the lower-right region, indicating fewer transitions even on complex instances.

Table~\ref{tbl:results} complements this figure by reporting the number of instances successfully solved—i.e., for which a correct controller was synthesized within the given budget—by each policy (RA, RL, GCRL) across domains.
Notably, GCRL solves more large-scale instances than both RA and RL in all domains except DP, indicating superior generalization performance.

We now dive into each scenario and describe the trends observed in the results.
In \textbf{[AT]}, some RL runs succeed in synthesizing controllers with minimal transitions, indicating that reinforcement learning can discover efficient paths. However, this behavior is unstable, whereas GCRL demonstrates more consistent generalization performance, achieving fewer transitions on average and solving more instances.
In \textbf{[BW]} and \textbf{[TA]}, GCRL significantly outperforms both baselines in terms of transition efficiency, highlighting its ability to reuse structural exploration strategies learned during training.
In \textbf{[TL]}, while RA achieves low transition counts for smaller problem sizes, it fails to synthesize controllers for larger instances. GCRL, by contrast, scales successfully and solves these larger instances, resulting in the highest number of solved cases overall.
In \textbf{[DP]}, RA performs best in terms of both the number of solved instances and the minimal transitions required. This trend is consistent with prior work~\cite{Delgado_Sanchez}, where both their training results (Figure~1 in~\cite{Delgado_Sanchez}) and ours (Figure~\ref{fig:training}) show that learned policies consistently require more transitions than RA. This suggests that learning-based approaches may struggle to match handcrafted heuristics in structurally uniform domains.

\rev{
We note that RA sometimes outperforms RL mainly due to our limited training budget.
While Delgado and Sánchez trained for $10^5$--$10^6$ steps per domain, we used only 100 episodes.
This small-budget setting was chosen to emphasize the sample efficiency of GCRL.
By introducing graph-based contextualization, GCRL propagates value information across related states and actions, achieving faster generalization in the early learning phase.
Evaluating this phase highlights GCRL’s key advantage, as long-horizon training is computationally prohibitive for graph-based models.
}

\subsection{Discussion}

\subsubsection{Heuristic Patterns Similar to RA}

Based on our analysis of the actual policy learned by GCRL, a key strength of the approach lies in its emergent---though not explicitly encoded---alignment with heuristic principles originally defined in RA.

In particular, RA employs a simple yet effective three-level prioritization scheme, as defined in Definition~13 of~\cite{CiolekDCS}: (1) uncontrollable transitions are always prioritized over controllable ones; (2) among uncontrollable transitions, those with unknown goal-distance ($d_j = \infty$) are preferred; and (3) otherwise, transitions with smaller estimated goal-distance $d_j$ are favored. This fixed strategy aggressively explores high-risk, unexplored transitions---especially uncontrollable ones---allowing RA to expose deadlocks or losing branches at an early stage.

Interestingly, although GCRL is not explicitly programmed with these rules, we observe that it frequently learns behaviors that resemble RA’s logic. Through reinforcement learning over graph-structured exploration histories, GCRL acquires an exploration policy that mirrors RA in the following ways. First, it demonstrates \emph{marked-state proximity}: transitions with smaller estimated distances $d_j$ to marked states are preferred (RA, Definition~11), allowing GCRL to prioritize frontiers likely to contribute to the synthesis of a winning controller. Second, it captures \emph{uncontrollable path risk}: according to RA's comparison rules (Definition~13), uncontrollable transitions---especially those with $d_j = \infty$ or $m_j = 2$---are prioritized as they may indicate unavoidable blocking behaviors. GCRL similarly learns to explore such transitions early, enabling it to identify losing branches quickly. Third, it engages in \emph{causal dependency reasoning}: RA models event enablement relationships using gap and dependency edges (Definitions~9--10). By encoding these structural relationships in the input graph, GCRL learns to infer that certain transitions (e.g., $\ell'$) cannot be meaningfully explored until their enabling transitions (e.g., $\ell$) have been expanded.
For example, in TL, uncontrollable loops that fail to reach marked states often trigger error propagation procedures such as \texttt{findNewErrorsIn} and \texttt{propagateError}. Experimental results show that GCRL assigns lower Q-values to such structures, thereby avoiding costly backtracking. Conversely, it tends to prioritize controllable transitions that lead toward successful completions, demonstrating the reuse of effective exploration patterns across different problem instances.

\subsubsection{Scalability via Structural Pattern Reuse}

Another equally critical factor enabling GCRL’s zero-shot generalization is its ability to reuse learned structural patterns across different scales. These patterns include controllable–uncontrollable branching structures, loops, and transitions toward marked states.

By encoding the explored LTS—including the frontier—into a graph, we observed that GCRL enables the following: (1) recognition of recurring local topologies—for example, in TL, machine-buffer pairs frequently form partial RA with uncontrollable transitions and loops, which GCRL identifies as early targets or avoidance points based on prior training; (2) transfer of useful abstractions—once GCRL learns to favor transitions leading toward marked states in small TL instances (e.g., $(n,k) = (2,2)$), it applies the same preference in larger instances (e.g., $(5,2)$) without requiring retraining; and (3) localized decision-making through $k$-hop subgraphs—decisions are based only on the frontier and its $k$-hop neighborhood, ensuring scalability even as the global graph size increases.

\subsubsection{Performance in Dining Philosophers}

As discussed in our experimental results, our method performs less effectively than RA in the Dining Philosophers (DP) domain—a trend that has also been observed in original RL approaches. We attribute this primarily to the nature of the DP domain, which is characterized by high symmetry and strictly local interactions. Each philosopher repeatedly requests and releases adjacent forks through nearly identical state machines, resulting in highly regular transition patterns. This structural regularity diminishes the effectiveness of learning-based exploration policies.

In such settings, the static heuristic used in Ready Abstraction (RA)—based on the tuple $\langle m_j, d_j \rangle$, where $m_j$ is a marker indicator and $d_j$ is the distance to the goal—assigns nearly uniform scores to all transitions. However, this uniformity is not problematic for RA, as its decision-making does not rely on fine-grained differences in $\langle m_j, d_j \rangle$ values. Instead, RA uses a fixed three-level priority scheme (Definition~13 of~\cite{CiolekDCS}), which aggressively prioritizes the exploration of uncontrollable transitions, especially those not yet evaluated. This behavior tends to quickly uncover deadlock-prone loops—for example, scenarios in which forks are taken but never released. In the DP domain, such loops are a common source of losing conditions, meaning RA’s policy naturally aligns with the key bottlenecks of the system.

By contrast, learning-based methods like RL and GCRL rely on reward signals to infer an optimal exploration strategy. Although GCRL benefits from graph-structured encoding of exploration history, the structural uniformity in DP—where nearly all components behave identically—reduces the discriminative power of its learned embeddings. As a result, the Q-value landscape becomes relatively flat, particularly when GCRL is trained only on small-scale DP instances such as $(n,k) = (2,2)$.

Therefore, RA’s superior performance in the DP domain should not be misinterpreted as evidence of its general superiority. Rather, it reflects a structural alignment: RA’s static prioritization scheme happens to coincide with the most informative transitions in DP, owing to the domain’s homogeneity and localized interactions. This coincidence allows RA to perform remarkably well despite its simplicity, while learning-based approaches are inherently disadvantaged by the lack of structural diversity in training signals.

\subsubsection{Considerations for Practical Use}
The performance breakdown in DP highlights that static heuristics like RA perform well in structurally uniform domains where uncontrollable transitions dominate. In such cases, RA's distance-based prioritization aligns naturally with the control objective, and learning-based methods offer limited additional advantage. In contrast, GCRL proves more effective in domains with structural irregularities, a mix of controllable and uncontrollable transitions, or varying event roles across components. It can adapt to problem-specific patterns that static heuristics may fail to capture.

For simple and uniform systems, RA remains a strong and efficient choice. However, in more complex or heterogeneous domains, GCRL is likely to yield better results. Therefore, understanding the structural characteristics of the target domain is essential when choosing between RA and GCRL for practical synthesis applications.

\section{Related Work}
\label{sec:related}

Reducing the computational cost of discrete controller synthesis has been extensively studied in the literature. Existing approaches can be broadly classified into the following categories.

The first line of research leverages state abstraction techniques to reduce the state space during the game space construction phase of discrete controller synthesis. A foundational method is presented in \cite{FLTL}, which reduces the state space by abstracting all potential requirement violations into a single error state during the parallel composition of the system and tester models.
\cite{jialong_RE22} extends this idea by introducing a multi-granularity abstraction strategy, enabling the construction of abstract models at varying levels of detail. This allows the synthesis process to adaptively select an appropriate abstraction level according to the specific requirements.
Building upon this abstraction paradigm, \cite{Yamauchi2024CDCS} further reduces computational costs by entirely avoiding the construction of states that violate safety requirements during game space generation.
Another notable direction is Stepwise Partial Synthesis \cite{stepwise}, which combines abstraction with incremental construction. This method incrementally builds only the minimal necessary game model for each individual requirement and performs localized analysis. As a result, it avoids constructing unnecessary portions of the domain model, significantly reducing overhead.

The second prominent approach is Directed Controller Synthesis, to which this work belongs. This technique incrementally explores the reachable, non-violating part of the state space starting from the initial state. Once a valid controller that satisfies all requirements is found, exploration terminates. This strategy avoids exhaustive exploration and thus significantly reduces computational cost.
To further improve exploration efficiency, several studies have introduced advanced heuristics. For example, \cite{CiolekDCS} explores the interdependencies between system components to narrow the search space and synthesize safe and non-blocking “directors.” \cite{Delgado_Sanchez} formulates the synthesis task as a reinforcement learning problem, using Deep Q-Networks (DQN) to learn optimal exploration strategies.
Building on this, the present work improves learning performance by integrating domain-specific structural information unique to the DCS context into the RL-based framework.

In addition to these major approaches, several alternative strategies have been proposed.
One such example is \cite{Hoffmann2020}, which advocates for planning with automata-network languages from model checking, enabling expressive controller synthesis by leveraging the structural strengths of temporal logic and automata representations.
For instance, \cite{jialong_EUC20} addresses the challenge of re-synthesis under environmental changes by proposing an efficient differential analysis algorithm that focuses only on the state space affected by such changes.
\cite{LazyCS} introduces a lazy synthesis technique in which the abstraction is computed on demand, rather than precomputing the full abstraction prior to controller synthesis. This allows for a more scalable synthesis process under safety and reachability objectives.
\cite{NotLosing} shifts focus from computing winning strategies to non-losing strategies within a 3-valued game framework. By reducing a 3-valued model checking game into two classical 2-valued parity games, it improves computational efficiency while maintaining correctness under partial information.

\section{Conclusion and Future Work}
\label{sec:conclusion}

In this paper, we addressed a key challenge in Directed Controller Synthesis (DCS): improving the efficiency of state-space exploration. We observed that existing methods—whether based on handcrafted heuristics or standard reinforcement learning—suffer from a form of contextual blindness, as they fail to incorporate information from the exploration history. To overcome this limitation, we introduced Graph Contextual Reinforcement Learning (GCRL), a framework that integrates Graph Neural Networks (GNNs) with Reinforcement Learning. By encoding the explored portion of the state space as a graph and analyzing it with a GNN, GCRL endows the learning agent with a form of structural memory, enabling it to make more informed decisions based on a holistic view of the explored region, rather than relying solely on local, short-sighted features.

Experiments across five benchmark domains demonstrate that GCRL not only accelerates learning but also scales better to larger, unseen problem instances. It consistently outperformed baseline methods in four domains by reducing training expansions and increasing synthesis success rates. While static heuristics like RA remained competitive in the highly symmetric domain like DP, GCRL proved especially effective in domains with rich and irregular structure.

In future work, there are several promising directions to explore. First, our current approach treats the exploration history as a static graph at each decision point, while the underlying process is inherently dynamic. 
We plan to incorporate Temporal Graph Networks (TGNs) to explicitly model the evolution of the graph over time, potentially capturing more nuanced temporal patterns and enhancing policy performance. 
Second, we aim to investigate more advanced graph learning techniques and develop more efficient methods for selecting relevant subgraphs to further improve computational efficiency and learning quality. 
\rev{Finally, we will address the memory efficiency of GCRL to improve its scalability. Although GCRL demonstrated strong performance, its graph construction can be memory-intensive in extremely large DES domains (e.g., CM). We plan to investigate methods like dynamic subgraph sampling and incremental graph encoding to substantially reduce the memory footprint, allowing the approach to scale to larger systems that are currently beyond its scope.}

\begin{appendices}
\label{sec:appendix_features}

\section*{Appendix: Feature Details for GCRL Model}

Table~\ref{tab:feature_details} provides a comprehensive description of the node and edge features used to construct the graph encoding for our GCRL model. These features are designed to capture both the local state properties and the broader structural context of the exploration history, informing the policy learned by the GNN.

\begin{table*}[!htb]
\centering
\caption{Detailed Description of Node and Edge Features for the GNN Model.
    \rev{following prior work (Table 1, RL method~\cite{Delgado_Sanchez}).}
}
\label{tab:feature_details}
\small
\resizebox{\textwidth}{!}{%
\begin{tabular}{lllp{6.5cm}}
\toprule
\textbf{Category} & \textbf{Feature Name} & \textbf{Type / possible values} & \textbf{Description} \\
\midrule
\multirow{5}{*}{\parbox{2.5cm}{\centering\textbf{Node \\ Features}}} 
 & Just explored flag & Boolean (\rev{2}) & A flag that is true if the node was discovered in the most recent exploration step. \\
 & Explored ratio & Real (\rev{continuous}) & The ratio of explored outgoing transitions to the total known outgoing transitions from this node. \\
 & Uncontrollable flag & Boolean (\rev{2}) & A flag that is true if the node has at least one outgoing uncontrollable transition. \\
 & Marked flag & Boolean (\rev{2}) & A flag that is true if the state is a marked state in the LTS specification. \\
 & Exploration context/phase & Categorical (3) & Flags indicating global synthesis progress (e.g., a marked state found, a winning state set). \\
\midrule
\multirow{9}{*}{\parbox{2.5cm}{\centering\textbf{Edge \\ Features}}} 
 & Event label & One-hot ($|A_E|$) & The event label of the transition, encoded as a one-hot vector. \\
 & Path labels (summarized) & Multi-hot ($|A_E|$) & A summary of event labels on the path to the source node, as a multi-hot vector. \\
 & Controllable flag & Boolean (\rev{2}) & A flag that is true if the transition's event is controllable. \\
 & Leads to marked state flag & Boolean (\rev{2}) & A flag that is true if the transition's target node is a marked state. \\
 & Synthesis phase info & Categorical (3) & Reflects the global synthesis phase at the time of decision, mirroring the corresponding node feature. \\
 & Child state classification & Categorical (4) & The current DCS classification of the target node (e.g., Win, Loss, None, Unexplored). \\
 & Child has uncontrollable flag & Boolean (\rev{2}) & A flag that is true if the target node has an outgoing uncontrollable transition. \\
 & Child explored status & Boolean (\rev{2}) & A flag that is true if any transition from the target node has already been explored. \\
 & Source is last expanded flag & Boolean (\rev{2}) & A flag that is true if the source node was the origin of the last exploration action. \\
\bottomrule
\end{tabular}%
}
\end{table*}

\end{appendices}

\bibliography{ref}

\end{document}